\documentclass[10pt,twocolumn,letterpaper]{article}

\usepackage{wacv}
\usepackage{times}
\usepackage{epsfig}
\usepackage{graphicx}
\usepackage{amsmath}
\usepackage{amssymb}
\usepackage[dvipsnames]{xcolor}

\def\wacvPaperID{828} 

\wacvfinalcopy 

\ifwacvfinal
\def\assignedStartPage{1} 
\fi

\ifwacvfinal
\usepackage[breaklinks=true,bookmarks=false]{hyperref}
\else
\usepackage[pagebackref=true,breaklinks=true,colorlinks,bookmarks=false]{hyperref}
\fi

\ifwacvfinal
\else
\fi

\begin{document}

\title{\textcolor{red}{Su}\textcolor{BurntOrange}{P}\textcolor{ForestGreen}{E}\textcolor{blue}{r}-\textcolor{Aquamarine}{S}\textcolor{Plum}{A}\textcolor{VioletRed}{M}: Using the \textcolor{red}{Su}pervision Signal from a \textcolor{BurntOrange}{P}ose \textcolor{ForestGreen}{E}stimato\textcolor{blue}{r} to Train a \textcolor{Aquamarine}{S}patial \textcolor{Plum}{A}ttention \textcolor{VioletRed}{M}odule for Personal Protective Equipment Recognition}

\author{Adrian \c{S}andru$^{1,2}$, Georgian-Emilian Du\c{t}\u{a}$^{1,2}$, Mariana-Iuliana Georgescu$^{1,2}$, Radu Tudor Ionescu$^{1,2}$, \\
$^1$SecurifAI, $^2$University of Bucharest\\
}

\maketitle

\begin{abstract}
We propose a deep learning method to automatically detect personal protective equipment (PPE), such as helmets, surgical masks, reflective vests, boots and so on, in images of people. Typical approaches for PPE detection based on deep learning are $(i)$ to train an object detector for items such as those listed above or $(ii)$ to train a person detector and a classifier that takes the bounding boxes predicted by the detector and discriminates between people wearing and people not wearing the corresponding PPE items. We propose a novel and accurate approach that uses three components: a person detector, a body pose estimator and a classifier. Our novelty consists in using the pose estimator only at training time, to improve the prediction performance of the classifier. We modify the neural architecture of the classifier by adding a spatial attention mechanism, which is trained using supervision signal from the pose estimator. In this way, the classifier learns to focus on PPE items, using knowledge from the pose estimator with almost no computational overhead during inference.
\end{abstract}

\setlength{\abovedisplayskip}{4pt}
\setlength{\belowdisplayskip}{4pt}

\vspace{-0.3cm}
\section{Introduction}

In many workplace environments, such as construction sites or oil and gas factories, there are potentially hazardous materials, dangerous chemicals or dangerous equipment (large moving parts, operating cranes), that present health or safety risks for the workers. In order to protect workers against accidents caused in hazardous or dangerous areas, the workers may be required to wear personal protective equipment (PPE). The personal protective equipment may include, but is not limited to, helmets, masks, glasses, uniforms, safety vests, boots, gloves, etc. In many countries, there are governmental regulations that require certain employees to be trained in the proper and effective use of PPE in hazardous or dangerous working environments. Nevertheless, in many cases, the workers choose not to wear PPE items because these items are uncomfortable, potentially causing sweat or skin irritations, or because these items prevent the workers from executing their tasks with precision, for example screwing or unscrewing a small screw with gloves on. Managing such risks is of utter importance to employers that aim at preventing unnecessary health or safety accidents. In this context, a system, such as the one proposed in this work, to automatically monitor the workers and verify if they are permanently wearing the mandatory personal protective equipment is extremely useful. We also recognize a broader application of our system in the context of the COVID-19 pandemic, which forced governments to impose new regulations requiring people to wear surgical masks in public places in order to reduce the spread of the SARS-CoV-2 virus. Our system could also be used to check for compliance with such regulations, as shown in our experiments.

Typical deep learning approaches for PPE detection are $(i)$ to train an object detector \cite{Fang-AC-2018,Kumar-ICMLIP-2020,Wang-Sensors-2020,Wu-AC-2019,Xie-PIC-2018,Zhafran-IES-2019}, resulting in a single-stage pipeline, or $(ii)$ to train a person detector and a classifier that takes the bounding boxes predicted by the detector and discriminates between people wearing and people not wearing the corresponding PPE items \cite{Chairat-WACV-2020,Dasgupta-CICT-2019,Nath-AC-2020}, resulting in a two-stage pipeline. A recent work \cite{Nath-AC-2020} demonstrated that the latter methodology produces more accurate results, mainly due to the poor ability of object detectors of detecting small objects. Another advantage of the two-stage approach is that it eliminates the necessity to annotate PPE items with bounding boxes. Given an accurate pre-trained person detector, we only need binary labels for each detected person with respect to each PPE item. This significantly reduces the annotation labor, allowing us to easily construct large data sets typically required for training deep learning models.

\begin{figure*}
\centering
\includegraphics[width=1.0\linewidth]{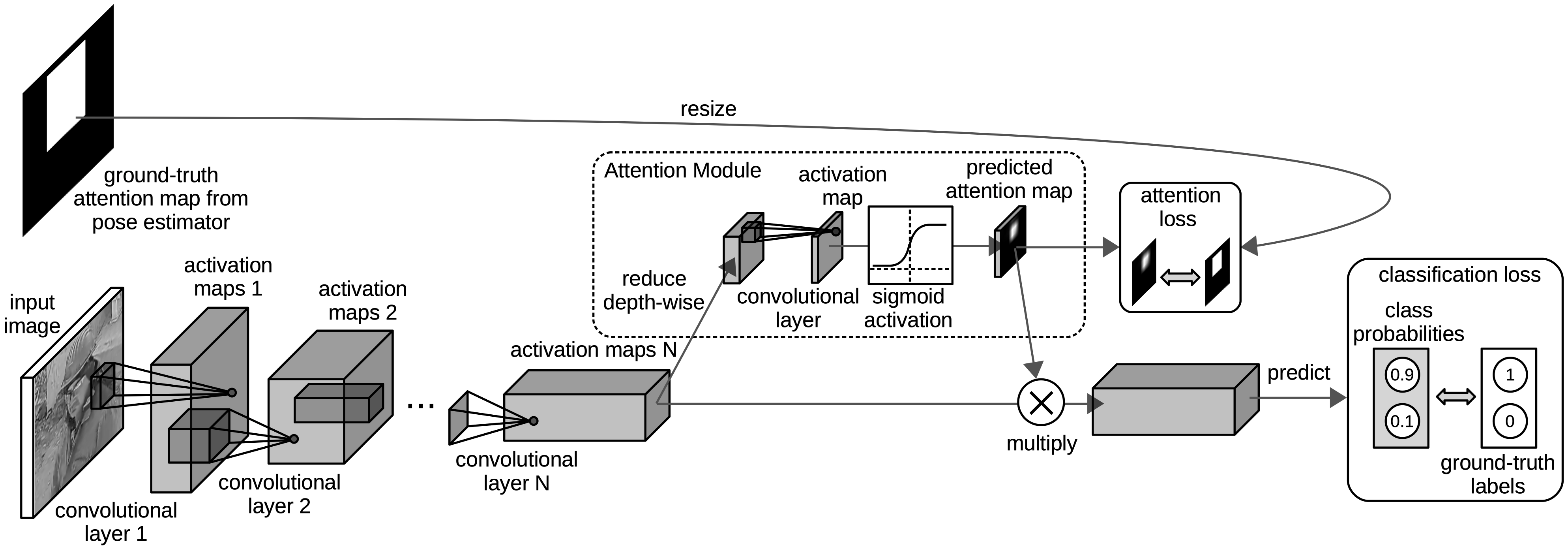}
\caption{A neural network classifier based on SuPEr-SAM, trained with classification loss and attention loss to discriminate people wearing helmets from people not wearing helmets. The supervision signal for the attention loss is provided by an articulated body pose estimator.}
\label{fig_1}
\vspace{-0.3cm}
\end{figure*}

We propose a novel and accurate approach that uses three components: a person detector, an articulated body pose estimator and a classifier. Our novelty consists in employing the pose estimator only at training time, providing pseudo-ground-truth attention maps to improve the performance of the classifier. We modify the neural architecture of the classifier by adding a spatial attention mechanism (SAM), which is trained using \textbf{su}pervision signal from the \textbf{p}ose \textbf{e}stimato\textbf{r} (SuPEr), as shown in Figure~\ref{fig_1}. In this way, the classifier learns to focus on PPE items, using knowledge from the pose estimator. At inference time, the pose estimator is removed from our processing pipeline, since it is only required during training to provide supervision to the attention module. Hence, the only computational overhead during inference is due to the spatial attention module. We term our approach SuPEr-SAM.

We conduct experiments on three data sets, comparing our approach with two baselines, namely the standard two-stage approach and a two-stage approach that integrates a standard spatial attention module. Each data set includes people wearing a certain PPE item type, namely helmets, masks or boots. We show that our approach significantly outperforms both baselines, irrespective of the PPE type.

The rest of this paper is organized as follows. We present the related works on PPE detection in Section~\ref{sec_related}. Our approach based on spatial attention trained with supervision signal from a pose estimator is described in Section~\ref{sec_method}. The comparative experiments are presented in Section~\ref{sec_experiments}. Finally, we draw our conclusion in Section~\ref{sec_conclusion}.

\vspace{-0.1cm}
\section{Related Work}
\label{sec_related}
\vspace{-0.1cm}

Some of the first systems developed for PPE detection \cite{Barro-CC-2012} are based on radio-frequency identification (RFID). Workers have to carry RFID devices or similar sensing devices which are programmed to identify the equipment based on proximity. For example, Barro-Torres \etal~\cite{Barro-CC-2012} presented an architecture formed of a wireless local area network and a body area network. 
Workers carry a microcontroller-based device that detects the presence of the PPE, sending reports to a central unit where alerts are generated. We note that such systems fail if the workers decide not to carry the RFID sensors, just as they would do with the personal protective equipment. Another problem with systems based on sensing devices is that they are costly to implement, since every piece of protective equipment needs to be tagged with a transmitting device. Unlike such systems, our approach employs a computer vision system that detects and recognizes PPE items in images captured by surveillance cameras. Since one surveillance camera can cover an area with multiple workers, a system based on image analysis is more cost effective than a system based on RFID sensors. Furthermore, the workers cannot trick the system by not wearing the sensing devices. Other than being applied for the same task, our method has no connection to works such as \cite{Barro-CC-2012}.

Before deep learning~\cite{LeCun-Nature-2015} was widely-adopted outside the computer vision community, researchers developed systems for PPE detection based on conventional machine learning or statistical methods \cite{Dahiya-IJCNN-2016,Silva-SIBGRAPI-2014,Li-ICACI-2017,Mneymneh-CCC-2017,Mneymneh-JCCE-2019,Moohialdin-EASEC-2019,Mosberger-Sensors-2014,Mosberger-ICRA-2015,Shrestha-JCE-2015,Wu-CI-2018}. It seems that the majority of systems, even from the recent literature~
\cite{Li-ICACI-2017,Mneymneh-CCC-2017,Mneymneh-JCCE-2019,Moohialdin-EASEC-2019,Wu-CI-2018}, still rely on handcrafted methods. In order to detect helmets, Shrestha \etal \cite{Shrestha-JCE-2015} employed a standard face detector. The detected faces are passed to a contour extraction method, which checks that the contour forms a semi-circle to confirm the presence of a helmet. The method is not reliable under low-resolution or blurry images or when the faces or helmets are partially occluded. Dahiya \etal \cite{Dahiya-IJCNN-2016} compared three methods based on handcrafted feature extraction: Histogram of Oriented Gradients (HOG), Scale-Invariant Feature Transform (SIFT) and Local Binary Patterns (LBP). Their method is designed to detect bike or motorbike riders without helmet. Wu \etal \cite{Wu-CI-2018} and Mneymneh \etal \cite{Mneymneh-CCC-2017} also employed a set of handcrafted image descriptors, which are combined together instead of being used independently. Nevertheless, these methods are tested on very few images, making it hard to draw generic conclusions. Moohialdin \etal \cite{Moohialdin-EASEC-2019} employed decision trees for PPE and posture detection. Decision trees are part of the standard machine learning methods, which cannot model the complexity of data affected by large variations. Li \etal \cite{Li-ICACI-2017} used a background subtraction algorithm to detect people. Their system is not reliable when objects other than people enter the analyzed scene, because these objects belong to the foreground. 
Another system based on background subtraction and handcrafted features is proposed in \cite{Mneymneh-JCCE-2019}. The system is trained and tested on a few hundred images collected in ideal conditions, which raises questions about the robustness of the system with respect to illumination changes, pose variation and occlusion. 
Mosberger \etal \cite{Mosberger-Sensors-2014} handcrafted a method to detect reflective safety garments. The method is specifically designed to detect the reflective patterns using thresholding and contour extraction. This limitation makes the method inapplicable to other PPE items that are not reflective, \eg glasses or masks. Moreover, their system requires an expensive camera mount, thus being less cost effective than our approach. 
Mosberger \etal \cite{Mosberger-ICRA-2015} enhanced their previous method by employing multi-band Hough Forests. While being more accurate than their former system \cite{Mosberger-Sensors-2014}, their latter system \cite{Mosberger-ICRA-2015} suffers from the same limitations indicated above.

We note that, in general, the systems based on conventional machine learning methods, \eg Support Vector Machines \cite{Li-ICACI-2017} or decision trees \cite{Moohialdin-EASEC-2019}, and handcrafted feature extraction, \eg HOG \cite{Silva-SIBGRAPI-2014,Mneymneh-CCC-2017,Wu-CI-2018}, SIFT \cite{Dahiya-IJCNN-2016} or LBP \cite{Dahiya-IJCNN-2016,Mneymneh-CCC-2017,Wu-CI-2018}, exhibit poor modeling capacity. Such systems do not cope well under illumination changes, aspect variations or occlusion, among other factors of variation in images. To this end, researchers and engineers turned their attention to deep learning methods, which provide superior results due to their capacity to learn a hierarchy of features from large amounts of data \cite{LeCun-Nature-2015}. Our system is based on deep learning, thus being able to achieve much better accuracy rates than systems based on conventional machine learning or handcrafted features. Our method is fundamentally different from \cite{Dahiya-IJCNN-2016,Silva-SIBGRAPI-2014,Li-ICACI-2017,Mneymneh-CCC-2017,Mneymneh-JCCE-2019,Moohialdin-EASEC-2019,Mosberger-Sensors-2014,Mosberger-ICRA-2015,Shrestha-JCE-2015,Wu-CI-2018}, since, in our case, the features are automatically learned instead of being designed by hand.

The PPE detection approaches based on deep learning are divided into two categories: $(i)$ training an object detector for PPE item detection or $(ii)$ training a person detector and a classifier that takes the bounding boxes predicted by the person detector as input. 
The majority of deep learning systems \cite{Fang-AC-2018,Kumar-ICMLIP-2020,Wang-Sensors-2020,Wu-AC-2019,Xie-PIC-2018,Zhafran-IES-2019} perform PPE detection in a single step, by training an object detector for the corresponding object classes. For example, Fang \etal \cite{Fang-AC-2018} trained a Faster R-CNN object detector \cite{Ren-NIPS-2015} on more than 100,000 images to detect people not wearing helmets. Another approach for helmet detection based on Faster R-CNN is presented in \cite{Zhafran-IES-2019}. In other works \cite{Kumar-ICMLIP-2020,Xie-PIC-2018}, the authors approached the task of helmet detection using the YOLO object detector \cite{Redmon-CVPR-2016}. Another approach based on a fast deep object detector is that of Wang \etal \cite{Wang-Sensors-2020}. For a fast processing time, their detector relies on a lightweight convolutional neural network as backbone. 
Wu \etal \cite{Wu-AC-2019} proposed a deep object detector based on multi-level features to detect people wearing helmets. To progressively encode the multi-level features, the authors employed reverse progressive attention. Their attention mechanism is trained in a conventional way. Unlike Wu \etal \cite{Wu-AC-2019}, we train the attention mechanism in a novel and different way. While they use attention in the detector, we employ attention in the classifier. Our novelty consists in training the attention maps by minimizing the mean absolute error with respect to a pseudo-ground-truth attention mask provided by a pose estimator. By using supervision from a pose estimator, our method is able to obtain better accuracy rates compared to attention mechanisms based on conventional training (without additional components added to the loss).

Methods that detect PPE items in a single stage \cite{Fang-AC-2018,Kumar-ICMLIP-2020,Wang-Sensors-2020,Wu-AC-2019,Xie-PIC-2018,Zhafran-IES-2019} are typically less accurate than two-stage methods performing people detection and PPE recognition \cite{Chairat-WACV-2020,Dasgupta-CICT-2019,Nath-AC-2020}. The main issue with methods based solely on object detectors is the inability of object detectors to detect small objects, \eg helmets or boots, from a far range. Instead of trying to detect these small objects relative to the entire scene, two-stage methods detect people, which are proportionally larger. For this reason, the latter approaches have superior accuracy rates. This statement is confirmed by the experimental results presented in \cite{Nath-AC-2020}. Our method is based on a two-stage approach, hence it is substantially different from methods performing detection in a single stage \cite{Fang-AC-2018,Kumar-ICMLIP-2020,Wang-Sensors-2020,Wu-AC-2019,Xie-PIC-2018,Zhafran-IES-2019}.
Some recent works \cite{Chairat-WACV-2020,Dasgupta-CICT-2019} focused on helmet detection of bike or motor riders. These systems use the YOLO object detector to detect people. Then, the authors trained convolutional neural networks to classify the detected people in two classes: violation (without helmet) or non-violation (with helmet). Different from such methods, our approach relies on an additional system, namely an articulated body pose estimator, that provides supervision to a spatial attention module inserted in the neural network classifier. This leads to accuracy improvements with an imperceptible computational time increase. To our knowledge, we are the first to train a spatial attention mechanism with supervision signal from a pose estimator.


\vspace{-0.05cm}
\section{Method}
\label{sec_method}
\vspace{-0.1cm}
\subsection{Overview of Processing Pipelines}
\vspace{-0.05cm}

The training and inference pipelines of the proposed method are slightly different, since the pose estimator is not employed during inference. We next present the steps involved in each of the two pipelines of our method.

\noindent
\textbf{Training pipeline.}
During training, the input images are processed by three neural models: a person detector, a pose estimator, one or more PPE classifiers. The number of classifiers is determined by the number of distinct types of PPE that need to be detected. We assume that the person detector and the body pose estimator are pre-trained models. We still need a set of labeled images to train our classifiers. First, we provide the training images to the person detector that outputs bounding boxes enclosing the detected people. Second, we apply the pose estimator on each image region (bounding box), providing as output a skeleton of the body indicating the location of body parts. From an entire bounding box, we automatically crop out a sub-region that can potentially contain a specific PPE item, \eg a helmet, and provide it to the corresponding classifier, \eg the helmet classifier. From the corresponding skeleton, we construct a pseudo-ground-truth attention mask indicating the location of the body part of interest inside the crop. For example, the body part of interest for the helmet classifier is the head. In summary, the input for a classifier is formed of an image, for example representing the upper body, and an attention mask, typically indicating the location of the body part of interest, for example the head. We process the input image through a series of convolutional blocks, as illustrated in Figure~\ref{fig_1}. We pass the activation maps from the last convolutional layer to a spatial attention module. We train each classifier by jointly minimizing an objective composed of a classification loss and an attention loss.

\noindent
\textbf{Inference pipeline.}
During inference, we apply the person detector on test images, obtaining a set of bounding boxes. From each bounding box, we crop out sub-regions that can potentially contain certain PPE items and provide them to the corresponding classifiers. We process each image crop through the corresponding convolutional neural network (CNN) classifier and its spatial attention module. The classifier provides a binary prediction for each input image, indicating whether the (partially-truncated) person in the image wears the corresponding PPE item or not. Additionally, the predicted attention map can be used to explain the model's decision for a given input image.

\subsection{Neural Models}

\noindent
\textbf{Person detector.} To detect people in images, we employ the YOLOv3 \cite{Redmon-arXiv-2018} object detector, which is pre-trained on the MS COCO data set \cite{Lin-ECCV-2014}. YOLOv3 \cite{Redmon-arXiv-2018} is a single-stage object detector that offers superior detection performance and high speed (more than $30$ frames per second on a single GPU). For each input frame, the object detector returns the bounding boxes of the objects together with the corresponding class labels. From these, we are only interested in the bounding boxes from the \emph{person} category.

\noindent
\textbf{Pose estimator.} To obtain the body pose of each detected person, we apply a pre-trained pose estimator based on Part Affinity Fields \cite{Cao-CVPR-2017}. The chosen pose estimator relies on global context, jointly learning part locations and their associations by using two branches in the same sequential estimation process.
For each input image, the pose estimator processes the image to recover the pose of the articulated bodies depicted in the image, representing the pose as a skeleton. The skeleton is a tree-structured chain with nodes representing body joints and edges representing rigid body segments. Each edge is associated with a specific body part, enabling us to build pseudo-ground-truth attention maps for the body parts of interest to the PPE classifiers. We note that the computational time of the pose estimator is not of utter importance, as it is used only at training time.

\noindent
\textbf{PPE classifier.}
Our PPE classifier is a lightweight CNN based on the MobileNetV2 architecture \cite{Sandler-CVPR-2018}, which is equivalent with AlexNet \cite{Krizhevsky-NIPS-2012} in terms of accuracy, yet offers faster processing times, thus being suitable even for mobile devices. In the standard MobileNetV2 model, which is pre-trained on ImageNet~\cite{Russakovsky-IJCV-2015}, we integrate SuPEr-SAM, a spatial attention module that receives supervision signal from the pose estimator.
As illustrated in Figure~\ref{fig_1}, the spatial attention module is introduced after the last convolutional layer. Since the activation maps from the last convolutional layer are passed to the spatial attention module, the neural network is forced to focus on the body part of interest with respect to the classification task. The proposed spatial attention module is composed of two depth-wise reduction layers. The goal of each reduction layer is to reduce the depth of the input tensor (a set of activation maps) to one, while preserving the spatial dimensions. Each reduction layer applies a different rule for the depth-wise reduction operation, \ie one selects the maximum value and the other computes the average. Each reduction layer produces a 2D activation map as output. The activation maps from the two reduction layers are stacked in the third dimension to form a 3D tensor. The tensor is then provided as input to a convolutional layer with a single $7\!\times\!7$ filter. Being formed of a single filter, the output of the convolutional layer is a single 2D activation map. The resulting activation map is passed through a sigmoid gate, scaling the values inside the  map to the interval $[0,1]$. The final activation map is interpreted as an attention mask denoted by $\hat{A}$. The attention mask is multiplied element-wise with each activation map that was provided as input to the attention module. The result is a set of activation maps with high filter responses only in the region indicated by the attention mask. To focus on the correct region, the attention mask $\hat{A}$ is trained in order to minimize an attention loss with respect to a pseudo-ground-truth attention mask $A$, which is automatically computed with the help of the pose estimator. Since each classifier's goal is to detect a specific kind of PPE covering a certain body part, the ground-truth activation mask is generated such that it contains a region of interest in the location where the pose estimator found the body part of interest. The region of interest is indicated as a rectangle filled with the value 1 in the ground-truth attention mask, the rest of the attention mask being filled with the value 0. Since the size of the ground-truth attention mask does not match the size of the predicted attention mask coming out of the spatial attention module, the ground-truth attention mask is rescaled such that the masks have the same size.

\subsection{Loss Function}

Let $X \in \mathbb{R}^{h \times w}$ be an input image of $h \times w$ pixels, \eg representing the upper body of a person, and $y \in \{0,1\}$ a ground-truth label associated to the input image. The classifier acts as a function $f:\mathbb{R}^{h \times w} \rightarrow [0,1]$  with parameters (weights) $\theta$, providing as output a class probability $\hat{y}$ indicating whether the input $X$ belongs to the class $y$. Then, the parameters $\theta$ are optimized such that the classifier $f$ is minimizing a classification loss function. In our case, the classification loss is the binary cross-entropy:
\begin{equation}
\mathcal{L}_{class}= - \sum_{i=1}^n y_i \cdot \log(\hat{y}_i) + (1 - y_i) \cdot \log(1-\hat{y}_i),    
\end{equation}
where $\hat{y}_i=f(\theta, X_i)$, $\mathcal{S}=\{(X_i,y_i) | X_i \in \mathbb{R}^{h \times w}, y_i \in \{0,1\}, \forall i=\overline{1,n} \}$ is a training set of images and $n$ is the size of the training set.
In addition to the classification loss function, our method optimizes the classifier towards minimizing an attention loss function. Since our attention mask $A$ contains binary values, we opt to train the attention module using binary cross-entropy:
\begin{equation}\label{eq_loss_attention}
\begin{split}
\mathcal{L}_{attention} = - \sum_{i=1}^n \sum_{j=1}^{h'} \sum_{k=1}^{w'} A_{i,j,k} \cdot &\log(\hat{A}_{i,j,k}) + \\
+ (1 - A_{i,j,k}) \cdot &\log(1-\hat{A}_{i,j,k}),
\end{split}
\end{equation}
where $n$ is the size of the training set, $h'$ and $w'$ are the height and the width of the predicted attention mask, $A_i$ is the ground-truth attention mask associated to the example $X_i$, $\hat{A}_i$ is the predicted attention mask for the example $X_i$, and $j$ and $k$ iterate through the components of the matrices $A_i$ and $\hat{A}_i$.

We combine the classification loss and the attention loss into a joint loss function expressed as follows:
\begin{equation}\label{eq_loss_joint}
\mathcal{L}_{joint} = \lambda \cdot \mathcal{L}_{class} + (1 - \lambda) \cdot \mathcal{L}_{attention},
\end{equation}
where $\lambda$ is a parameter between $0$ and $1$ that controls the balance between the classification loss and the attention loss. Each PPE classifier is trained using the Adam optimizer \cite{Kingma-ICLR-2014}, minimizing the joint loss function defined in Equation~\eqref{eq_loss_joint}. It is important to note that, during inference, the ground-truth attention masks are no longer needed, since these are used only at training time to compute the attention loss defined in Equation~\eqref{eq_loss_attention}. Therefore, the articulated body pose estimator is not required at inference time, significantly reducing the processing time of the proposed method. Since only the person detector and the classifiers are employed during inference, our entire architecture can process an input video stream in real-time at about $30$ frames per second ($32.6$ milliseconds per frame, with an average of one person per frame) using one GPU (see Table~\ref{tab_running_time}).

\vspace{-0.1cm}
\section{Experiments}
\label{sec_experiments}
\vspace{-0.1cm}
\subsection{Data Sets}
\vspace{-0.1cm}

We evaluate our model on three types of PPE, namely helmets, boots and surgical masks.

\noindent
\textbf{Helmets data set.} The helmets data set is composed of people wearing and people not wearing helmets. We collected a total of $13,\!481$ images of people wearing helmets and $4,\!170$ images of people not wearing helmets. We split the data set into $12,\!286$ examples for training, $2,\!628$ examples for validation and $2,\!737$ examples for testing, using stratified sampling. Hence, the ratio of people wearing helmets to people not wearing helmets is roughly $3\!:\!1$ in all three subsets.

\noindent
\textbf{Boots data set.} The boots data set contains images with people wearing and people not wearing protective boots. We gathered a total of $4,\!784$ images of people wearing protective boots and $988$ images of people wearing other types of shoes. We divide the data set in $3,\!988$ examples for training, $866$ examples for validation and $918$ examples for testing. Since we used stratified sampling for splitting the data, the ratio of people wearing protective boots to people not wearing them is nearly $4.5\!:\!1$ in all three subsets.

\noindent
\textbf{Surgical Masks data set.} The surgical masks data set is formed of images of people with and without masks. We collected a total of $1,\!535$ images of people wearing surgical masks and $5,\!473$ images of people that do not wear masks. Using stratified sampling, we split the data set in $3,\!421$ examples for training, $1,\!710$ examples for validation and $1,\!877$ examples for testing. The ratio of people wearing to people not wearing masks in each subset it is $1\!:\!3.5$.

\begin{table*}[!t]
\setlength\tabcolsep{1.6pt}
\begin{center}
\begin{tabular}{|l|l|c|c|c|c|c|c|c|}
\hline  
Data Set & Method  &  Accuracy         & \multicolumn{3}{|c|}{Macro} & \multicolumn{3}{|c|}{Weighted}  \\     
 
\cline{4-9}     &    &      &     Precision & Recall & F1-score & Precision & Recall & F1-score \\ 
\hline
\hline

  &  MobileNet &  $96.90\!\pm\!0.9$ &  $96.96\!\pm\!1.4$ &  $95.01\!\pm\!1.1$ & $95.93\!\pm\!1.2$ & $96.91\!\pm\!1.0$ & $96.90\!\pm\!0.9$ & $96.87\!\pm\!0.9$ \\ 
\cline{2-9}
Helmets   &  MobileNet+SAM & $95.76\!\pm\!1.0$ &  $96.53\!\pm\!0.5$ &  $92.56\!\pm\!2.2$ & $94.28\!\pm\!1.5$ & $95.88\!\pm\!0.9$ & $95.76\!\pm\!1.0$ & $95.65\!\pm\!1.1$ \\ 
\cline{2-9}
   &  MobileNet+SuPEr-SAM & $\mathbf{98.92}\!\pm\!0.4$ &  $\mathbf{98.89}\!\pm\!0.2$ &  $\mathbf{98.33}\!\pm\!0.9$ & $\mathbf{98.60}\!\pm\!0.5$ & $\mathbf{98.93}\!\pm\!0.4$ & $\mathbf{98.92}\!\pm\!0.4$ & $\mathbf{98.92}\!\pm\!0.4$ \\ 
\hline
\hline
  &  MobileNet &  $95.69\!\pm\!0.5$ &  $89.89\!\pm\!2.2$ &  $91.27\!\pm\!1.7$ & $90.47\!\pm\!0.8$ & $95.85\!\pm\!0.4$ & $95.69\!\pm\!0.5$ & $95.74\!\pm\!0.4$ \\ 
\cline{2-9}
Boots   &  MobileNet+SAM & $95.48\!\pm\!0.9$ &  $\mathbf{93.20}\!\pm\!2.1$ &  $85.69\!\pm\!3.7$ & $88.83\!\pm\!2.6$ & $95.37\!\pm\!0.9$ & $95.48\!\pm\!0.9$ & $95.26\!\pm\!1.0$ \\ 
\cline{2-9}
   &  MobileNet+SuPEr-SAM & $\mathbf{96.54}\!\pm\!0.2$ &  $91.48\!\pm\!1.1$ &  $\mathbf{92.93}\!\pm\!0.9$ & $\mathbf{92.17}\!\pm\!0.5$ & $\mathbf{96.63}\!\pm\!0.2$ & $\mathbf{96.54}\!\pm\!0.2$ & $\mathbf{96.57}\!\pm\!0.2$ \\ 
\hline
\hline
  &  MobileNet &  $93.90\!\pm\!1.1$ &  $88.59\!\pm\!1.6$ &  $95.69\!\pm\!0.6$ & $91.42\!\pm\!1.3$ & $95.16\!\pm\!0.6$ & $93.90\!\pm\!1.1$ & $94.17\!\pm\!1.0$ \\ 
\cline{2-9}
Masks   &  MobileNet+SAM & $94.74\!\pm\!1.1$ &  $89.88\!\pm\!1.8$ &  $\mathbf{96.23}\!\pm\!0.7$ & $92.50\!\pm\!1.5$ & $95.70\!\pm\!0.7$ & $94.74\!\pm\!1.1$ & $94.94\!\pm\!1.1$ \\ 
\cline{2-9}
   &  MobileNet+SuPEr-SAM & $\mathbf{95.26}\!\pm\!1.0$ &  $\mathbf{90.92}\!\pm\!1.9$ &  $96.08\!\pm\!0.5$ & $\mathbf{93.14}\!\pm\!1.4$ & $\mathbf{95.92}\!\pm\!0.7$ & $\mathbf{95.26}\!\pm\!1.0$ & $\mathbf{95.41}\!\pm\!1.0$ \\  
\hline
\end{tabular}
\end{center}
\vspace{-0.1cm}
\caption{Results of the MobileNet and the MobileNet+SAM baselines versus our MobileNet+SuPEr-SAM on the Helmets, the Boots and Surgical Masks data sets. Reported values are averaged over 5 runs. The best score on each data set and each metric is highlighted in bold.}
\label{tabel_results}
\vspace{-0.3cm}
\end{table*}

\vspace{-0.1cm}
\subsection{Evaluation Metrics}
\vspace{-0.1cm}

In the evaluation, we employ several metrics, such as the accuracy, the macro-averaged precision, the macro-averaged recall and the macro-averaged $F_1$ score. As the accuracy is equivalent to the micro-averaged $F_1$ score, we do not report micro-averaged metrics.
A macro-averaged metric computes the score for each class, then averages the results. For example, in the binary case, the macro-averaged precision is computed by calculating the precision for class $0$ and class $1$, then averaging the resulting scores. Since the macro-averaged metrics do not take into account the imbalanced class distributions in our data sets, we also report the weighted precision, the weighted recall and the weighted $F_1$ score. For instance, the weighted precision is computed by calculating the precision for class $0$ and class $1$, but instead of averaging the resulting scores, it weights the scores by the class probabilities. 

In the last experiment, we compare the two-stage approaches, including our own, with an approach that performs detection and recognition in a single step, considering the mean Average Precision (mAP). The mAP score is the mean of the AP scores corresponding to all object classes. The AP score for one class is represented by the area under the precision-recall (PR) curve. The PR curve is obtained by mapping each detected bounding box to the most-overlapping ground-truth bounding box, considering a minimum Intersection over Union (IoU) score of $0.5$~\cite{Everingham-IJCV-2015}.

\vspace{-0.1cm}
\subsection{Baselines and Models}
\vspace{-0.1cm}

\noindent
\textbf{MobileNet.} As our first baseline, we consider a two-stage approach that detects people using a pre-trained YOLOv3 model \cite{Redmon-arXiv-2018}, then classifies the cropped detections using a fine-tuned MobileNetV2 architecture \cite{Sandler-CVPR-2018}. 
 
\noindent
\textbf{MobileNet+SAM.} As our second baseline, we take into consideration a two-stage approach based on YOLOv3 and MobileNetV2, which introduces a spatial attention module (SAM) in the MobileNetV2 architecture. The module is placed immediately after the last convolutional layer.

\noindent
\textbf{MobileNet+SuPEr-SAM (ours).} Our model is most similar to the second baseline, the only difference being that the spatial attention module (SAM) is trained using \textbf{su}pervision signal from a \textbf{p}ose \textbf{e}stimato\textbf{r} (SuPEr). Our approach is therefore called SuPEr-SAM.

\noindent
\textbf{YOLO on faces.} Another relevant baseline is to fine-tune the YOLOv3 model to detect and recognize PPE items directly. Since this baseline requires manual bounding box labeling, we test it on a single data set, namely the Surgical Masks data set. 
Hence, the model detects and recognizes two object categories: faces with masks and faces without masks. Both categories are required, since we also need to know when people do not wear the mask.

\begin{figure}[!t]
\centering
\includegraphics[width=1.0\linewidth]{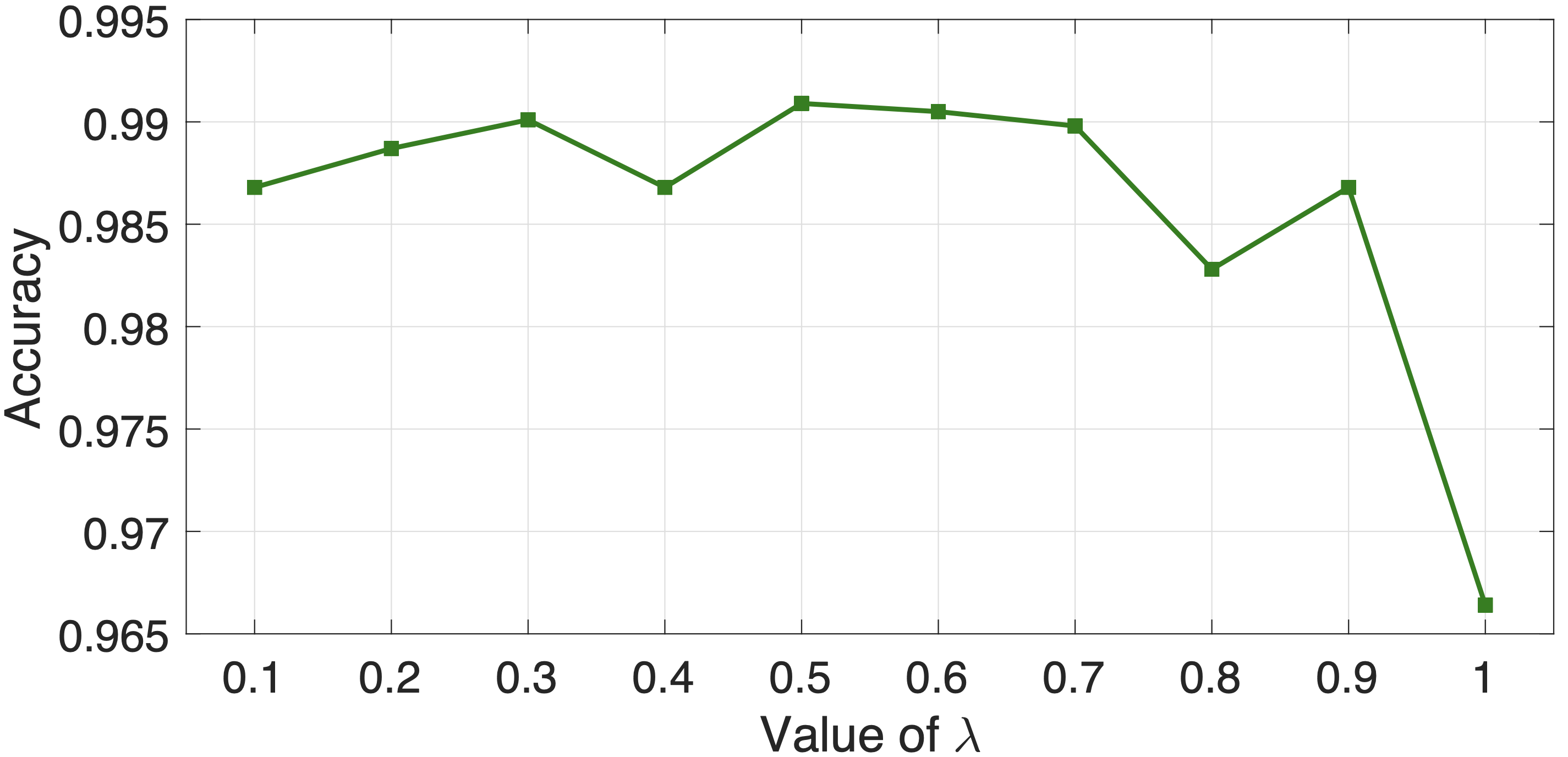}
\vspace{-0.4cm}
\caption{Validation accuracy rates on the Helmets data set, while tuning $\lambda$ in Equation~\eqref{eq_loss_joint} between $0.1$ and $1$.}
\label{fig_tuning_lambda}
\vspace{-0.3cm}
\end{figure}

\vspace{-0.1cm}
\subsection{Parameter and Implementation Choices}
\vspace{-0.1cm}

We trained the MobileNetV2 \cite{Sandler-CVPR-2018} classification models in Keras \cite{Chollet-keras-2015}. We employed the Adam optimizer \cite{Kingma-ICLR-2014} with the learning rate set to $10^{-4}$, while keeping the default values for the other parameters of Adam. Each model was trained for a maximum of 150 epochs using early stopping with respect to the validation set. In each experiment, we set the mini-batch size to $32$ samples. 
We tuned the parameter $\lambda$ in Equation~\eqref{eq_loss_joint} on each validation set, considering values between $0.1$ and $1$, using a step of $0.1$. We obtained optimal results using $\lambda=0.5$. In Figure~\ref{fig_tuning_lambda}, we illustrate the parameter tuning process on the Helmets data set.

\vspace{-0.1cm}
\subsection{Quantitative Results}
\vspace{-0.1cm}

We present the quantitative results on the three data sets in Table~\ref{tabel_results}. For each system, we report the average scores and the standard deviations for $5$ runs, in order to account for the variation that commonly appears in the initialization and the stochastic training process of neural networks.

\noindent
\textbf{Results on Helmets.}
On the Helmets data set, we outperform the two baselines, namely MobileNet \cite{Sandler-CVPR-2018} and MobileNet+SAM, by large margins, regardless of the metric. For instance, we surpass the standard MobileNet by $2.02\%$ in terms of the accuracy rate and by $2.67\%$ in terms of the macro-averaged $F_1$ score, respectively. Interestingly, the MobileNet+SAM model attains lower performance levels than the standard MobileNet. By adding SAM, we observe performance drops higher than $1\%$ for almost all metrics. This shows that the spatial attention module was not able to learn relevant spatial information without direct supervision. 
It is worth noting that our approach is more stable, having lower standard deviations over the 5 runs compared to the two baselines. For example, our standard deviation in terms of the macro-averaged precision is $0.2\%$, which is significantly lower than the standard deviation of the baseline MobileNet architecture, which is equal to $1.4\%$. In summary, the results clearly show the benefit of using the supervision signal from the pose estimator to train the spatial attention module.

\noindent
\textbf{Results on Boots.}
The results presented in Table~\ref{tabel_results} indicate that we generally outperform both baselines on the Boots data set. Our approach attains a macro-averaged precision of $91.48\%$, being surpassed by the MobileNet+SAM, which yields a macro-averaged precision of $93.20\%$. Even though SAM seems to perform better than SuPEr-SAM in terms of the macro-averaged precision, when we consider the macro-averaged recall, our method attains a score that is $7.24\%$ higher. Consequently, the macro-averaged $F_1$ score of MobileNet+SuPEr-SAM is superior. Moreover, we note that MobileNet+SAM attains generally lower performance scores, even compared to the standard MobileNet architecture. Once again, this proves that the integration of the standard spatial attention module is rather harmful. On the contrary, SuPEr-SAM brings performance gains, regardless of the evaluation metric. Our model is also more stable, having a maximum standard deviation of $1.1\%$, compared to $3.7\%$ obtained by MobileNet+SAM and $2.2\%$ obtained by MobileNet, respectively.

\begin{table}[!t]
\setlength\tabcolsep{1.6pt}
\begin{center}
\begin{tabular}{|l|c|c|c|}
\hline 
Method  &  No Mask & Mask & mAP \\ 
\hline    
\hline
YOLO on faces         & $91.93 \%$ & $83.90\%$ & $87.91\% $\\
\hline 
MobileNet               & $90.15 \%$ & $86.39\%$ & $88.27\% $\\
\hline 
MobileNet+SAM           & $90.33 \%$ & $86.18\%$ & $88.25\% $\\
\hline 
MobileNet+SuPEr-SAM     & $90.31 \%$ & $87.19\%$ & $88.75\% $\\
\hline 
\end{tabular}
\end{center}
\vspace*{-0.1cm}
\caption{AP and mAP scores of a single-stage pipeline (YOLO on faces) versus three two-stage pipelines based on YOLO on people and various MobileNet classifiers, with and without attention.}\label{tab_yolo}
\vspace*{-0.3cm}
\end{table}

\begin{figure*}[!t]
\centering
\includegraphics[width=0.92\linewidth]{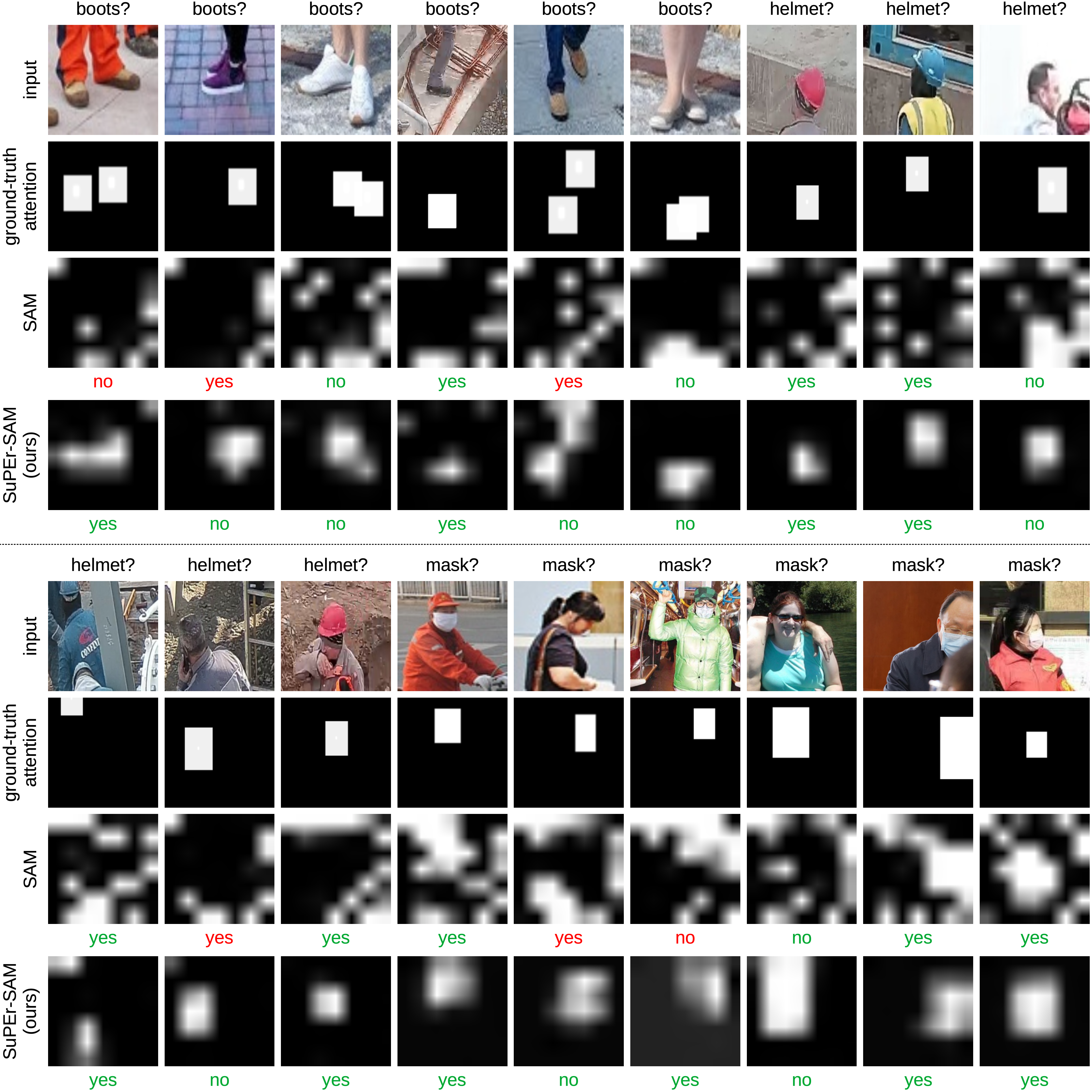}
\vspace{-0.1cm}
\caption{Examples of cropped input images and associated pseudo-ground-truth attention masks versus the attention masks predicted by SAM and SuPEr-SAM, respectively. The examples are chosen from the Boots, the Helmets and the Surgical Masks test sets. The labels predicted by MobileNet+SAM or by MobileNet+SuPEr-SAM are placed immediately below the corresponding attention masks. Correct labels are colored in green and wrong labels are colored in red. Best viewed in color.}
\label{fig_2}
\vspace{-0.35cm}
\end{figure*}

\noindent
\textbf{Results on Surgical Masks.}
The macro-averaged recall of MobileNet+SAM is slightly higher ($0.15\%$) than the macro-averaged recall obtained by MobileNet+SuPEr-SAM, but our macro-averaged precision is superior, with a difference of $1.04\%$. Consequently, in terms of the macro-averaged $F_1$ score (which combines precision and recall), we obtain an improvement of $0.64\%$ compared to MobileNet+SAM. We also surpass the standard MobileNet model, regardless of the evaluation metric. Hence, we conclude that our approach outperforms the two baselines on the Surgical Masks data set. We also note that, on this data set, MobileNet+SAM outperforms the standard model. This is the only case in which we observe performance gains due to the integration of standard spatial attention. 

\noindent
\textbf{One-stage versus two-stage pipelines.} 
In Table~\ref{tab_yolo}, we compare the results of the one-stage pipeline (YOLO on faces) with the results of the two-stage pipelines on the Surgical Masks data set. We present the results of the one-stage pipeline only on one data set, because this approach requires human labor to annotate images at the bounding box level. This represents a disadvantage compared to the two-stage approaches, which require only class label annotations. We want to underline, once again, that the two-stage pipelines, denoted by MobileNet, MobileNet+SAM and MobileNet+SuPEr-SAM, are based on detecting people with YOLOv3 and on classifying cropped body parts with MobileNetV2.
The Average Precision (AP) of YOLO on detecting faces with masks is $83.90\%$. When we employ MobileNet+SuPEr-SAM on the detected people, we obtain a much higher AP score ($87.19\%$) for the same class (people with mask). In general, the two-stage pipelines exhibit lower AP scores for people without masks, when we compare them to the single-stage pipeline. However, the mAP scores of the two-stage pipelines are always higher than the mAP score of the single-stage pipeline. Our results confirm the already known fact~\cite{Nath-AC-2020} that two-stage approaches outperform single-stage models.


\vspace{-0.1cm}
\subsection{Running Time}
\vspace{-0.1cm}

In Table~\ref{tab_running_time}, we compare our model to the baselines in terms of running time. Since MobileNet+SAM and MobileNet+SuPEr-SAM use the same spatial attention module, there is no difference in terms of running time between the two models. MobileNet requires $8.7$ milliseconds (ms) to infer the class for one image, while MobileNet+SAM and MobileNet+SuPEr-SAM require $9.3$ ms for the same task. Hence, the spatial attention module increases the running time by less than $1$ ms (about $6.5\%$). The YOLOv3 detector requires $23.3$ ms per image, being more computationally intensive than MobileNetV2. With an average of one person per image, our complete two-stage pipeline requires $23.3+9.3=32.6$ ms.

\begin{table}[!t]
\setlength\tabcolsep{1.6pt}
\begin{center}
\begin{tabular}{|l|c|c|c|}
\hline 
Method  &  Detector Time & Classifier Time \\ 
\hline   
\hline
YOLO on faces  &  $23.3$ ms & - \\
\hline     
MobileNet        &  $23.3$ ms & $8.7$ ms \\
\hline 
MobileNet+SAM    &   $23.3$ ms & $9.3$ ms\\
\hline 
MobileNet+SuPEr-SAM  &  $23.3$ ms  & $9.3$ ms \\
\hline 
\end{tabular}
\end{center}
\vspace*{-0.1cm}
\caption{Inference times in milliseconds (ms) per image of MobileNet+SuPEr-SAM versus three baselines. Reported times are measured on a machine with an Intel i9-9900K $3.6$GHz CPU, $64$GB of RAM and an NVIDIA GeForce RTX 2080Ti $11$GB GPU.}\label{tab_running_time}
\vspace*{-0.3cm}
\end{table}

\vspace{-0.1cm}
\subsection{Qualitative Results}
\vspace{-0.1cm}

Figure~\ref{fig_2} illustrates examples of cropped images that are provided to MobileNet+SAM and MobileNet+SuPEr-SAM during inference. For the selected examples, we computed the pseudo-ground-truth attention masks, using the pose estimator to determine the location of body parts. This enables us to compare the pseudo-ground-truth attention masks with the masks predicted by MobileNet+SAM and by MobileNet+SuPEr-SAM, respectively. In general, the attention masks predicted by SuPEr-SAM are significantly more correlated to the pseudo-ground-truth attention masks than the attention masks predicted by SAM. For SAM, we observe visual correlations for the sixth image on the first row (counting from left to right) and for the last two images on the second row. Although, in general, the attention masks produced by SAM seem uncorrelated with the locations of the PPE items, the predicted class labels are still correct in most cases. Nevertheless, we conclude that the illustrated examples clearly show that SuPERr-SAM helps the MobileNet classifier to focus on the right regions.

\vspace{-0.1cm}
\section{Conclusion}
\label{sec_conclusion}
\vspace{-0.1cm}

In this paper, we presented a two-stage approach for PPE recognition. Our contribution consists in introducing a pose estimator at training time, which provides a supervision signal to the spatial attention module incorporated in the classifier. Since the pose estimator is not required during inference, our pipeline adds a negligible margin (less than one millisecond) to the overall computational cost. The empirical results reported on three PPE data sets show that our contribution provides important accuracy gains. We conclude that our novel contribution is both effective and efficient.
In future work, we aim to extend our work beyond the task of PPE recognition, integrating SuPEr-SAM in other neural architectures. 

{\small
\bibliographystyle{ieee_fullname}
\bibliography{egbib}
}

\end{document}